%% file: acl.tex
\pdfoutput=1

\documentclass[11pt]{article}

\usepackage[final]{acl}

\usepackage{times}
\usepackage{latexsym}

\usepackage[T1]{fontenc}

\usepackage[utf8]{inputenc}

\usepackage{microtype}

\usepackage{inconsolata}

\usepackage{graphicx}

\usepackage{multirow}
\usepackage[table,xcdraw]{xcolor}
\usepackage{booktabs}       
\usepackage{amsfonts}       
\usepackage{nicefrac}       
\usepackage{microtype}      
\usepackage{xcolor}         
\usepackage{algorithm}
\usepackage{algorithmic}
\usepackage{amsmath}

\title{Ambiguity Awareness Optimization: Towards Semantic Disambiguation for Direct Preference Optimization}

\author{
 \textbf{Jian Li\textsuperscript{1, *}},
 \textbf{Shenglin Yin\textsuperscript{2, *}},
 \textbf{Yujia Zhang\textsuperscript{1, \dag, \ddag}},
 \textbf{Alan Zhao\textsuperscript{1, \dag}}, \\
 \textbf{Xi Chen\textsuperscript{1, \dag}},
 \textbf{Xiaohui Zhou\textsuperscript{1}},
 \textbf{Pengfei Xu\textsuperscript{1}} \\ 
 \\
 \textsuperscript{1}AI Technology Center of OVB, Tencent, China\\
 \textsuperscript{2}School of Computer Science, Peking University, China
}

\begin{document}
\maketitle
\footnotetext[1]{Equal contribution.}
\footnotetext[2]{Corresponding author.}
\footnotetext[3]{Project lead.}
\footnotetext[4]{Our code may be found at: \url{https://github.com/InsLin/AAO}.}
\input{sec/abs}

\input{sec/intro}
\input{sec/method}
\input{sec/exp}

\input{sec/related_work}
\input{sec/con}
\input{sec/limitations}

\bibliography{acl}

\end{document}

%% file: sec/abs.tex
\begin{abstract}
Direct Preference Optimization (DPO) is a widely used reinforcement learning from human feedback (RLHF) method across various domains. Recent research has increasingly focused on the role of token importance in improving DPO effectiveness. It is observed that identical or semantically similar content (defined as ambiguous content) frequently appears within the preference pairs. We hypothesize that the presence of ambiguous content during DPO training may introduce ambiguity, thereby limiting further improvements in alignment. Through mathematical analysis and proof-of-concept experiments, we reveal that ambiguous content may potentially introduce ambiguities, thereby degrading performance. To address this issue, we introduce Ambiguity Awareness Optimization (AAO), a simple yet effective approach that automatically re-weights ambiguous content to reduce ambiguities by calculating semantic similarity from preference pairs. Through extensive experiments, we demonstrate that AAO consistently and significantly surpasses state-of-the-art approaches in performance, without markedly increasing response length, across multiple model scales and widely adopted benchmark datasets, including AlpacaEval 2, MT-Bench, and Arena-Hard. Specifically, AAO outperforms DPO by up to 8.9 points on AlpacaEval 2 and achieves an improvement of by up to 15.0 points on Arena-Hard. 
\end{abstract}

%% file: sec/intro.tex
\section{Introduction}
\label{sec:intro}
To align large language models (LLMs) with human values and intentions, learning from human feedback is essential. \cite{ziegler2019fine,stiennon2020learning,bai2022training}. Reinforcement learning from human feedback (RLHF) has emerged as a key approach for aligning LLMs with human preferences and values \cite{christiano2017deep,ouyang2022training}. The core of this technique lies in the introduction of an explicit reward model during training to generate reward signals, and the application of reinforcement learning under a reference model that is consistent with the initial policy \cite{achiam2023gpt,touvron2023llama}. 

While RLHF enhances output quality, its reliance on multiple models and iterative sampling increases training complexity \cite{zhao2023slic,yuan2023rrhf,azar2024general,hong2024orpo,casper2023open}. To address this, direct alignment approaches like Direct Preference Optimization (DPO) \cite{rafailov2023direct} and its variants \cite{zhao2023slic,yuan2023rrhf,hong2024orpo,ethayarajh2024model,park2024disentangling,xu2024contrastive,ethayarajh2024kto,tang2024generalized,meng2024simpo} directly optimize LLMs based on human preferences, bypassing the need for separate reward models. These methods adjust the model’s loss by favoring preferred responses and penalizing dispreferred ones \cite{zhao2024rainbowpo,zeng2024token,liao2024tpo,liu2024tis}. However, DPO primarily emphasizes sequence-level preferences, overlooking the varying importance of individual tokens, which limits its effectiveness \cite{lin2024rho,zeng2024token,liu2024tis,gu2025mask}.

To investigate why distinguishing the importance of tokens is critical in DPO, we observe that many training pairs contain identical or semantically similar words (which we define as ambiguous contents). We hypothesize that training on ambiguous contents may lead to confusion, thereby limiting the performance of DPO. 

\begin{figure*}[t]
  \centering
  \includegraphics[trim=0 300 180 0, clip, width=1.\textwidth]{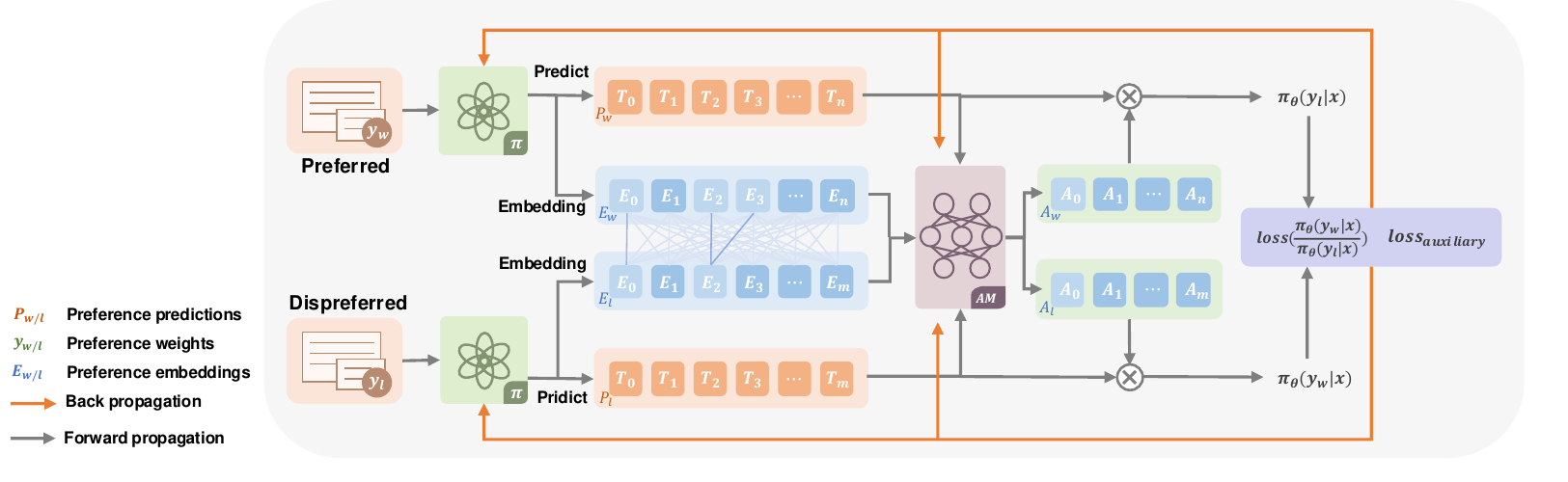}
  \caption{Diagram of our proposed AAO, in which background tokens among preference answers are re-weighted when computing cross-entropy loss. Firstly, AAO tokenizes preference pairs using a language model and encoding them into corresponding embeddings. Then AAO calculates the semantic similarity of the embeddings with cosine distance and decides the background tokens with a adaptive threshold. Finally, the background tokens are re-weighted during training.}
  \label{fig:pao_architecute}
  \vspace{-1em}
\end{figure*}

Building on this insight, we propose a simple yet effective method called Ambiguity Awareness Optimization (AAO) (shown in Figure \ref{fig:pao_architecute}), which can re-weight tokens to mitigate ambiguity during training by LLM itself. Specifically, we categorize response tokens into three types based on their semantic similarity (as measured by the LLM’s own embeddings), sorted from highest to lowest similarity: ambiguous tokens, transitional tokens, and key tokens. For each token type, we design distinct weight adjustment curves according to their semantic similarity. Furthermore, we introduce an adaptive module in the latent space that automatically determines the decision thresholds for the three token groups, enabling the model to dynamically decide these thresholds during training.

Extensive experiments were conducted on Llama3.1-8B \cite{grattafiori2024llama} and Mistral-7B \cite{jiang2024mixtral} models, including both base and instruction-tuned variants, utilizing widely adopted benchmark datasets such as AlpacaEval 2 \cite{li2023alpacaeval,dubois2024length}, MT-bench \cite{zheng2023judging}, and Arena-Hard \cite{li2024live}, among others. The results demonstrate that AAO consistently and significantly enhances the performance of DPO. 
Additionally, we further validated that AAO can alleviate the "squeeze effect" \cite{ren2024learning} present in DPO, revealing that the "squeeze effect" arises not only from sequence-level semantically similar pairs, but also from ambiguous tokens within the pairs. It is noteworthy that AAO does not rely on external models or additional data, making it highly flexible and easy to deploy.

Our work makes three key contributions:

\vspace{-0.5em}
\begin{itemize}
\itemsep=1pt
\item We conduct a thorough mathematical analysis of the effect of ambiguous content and perform comprehensive experiments, revealing the negative impact of ambiguous content on preference optimization.
\itemsep=1pt
\item We propose a simple yet effective method called AAO, which can automatically identify ambiguous content during training. Additionally, it can be seamlessly integrated with existing methods, enhancing their performance without much computational burden.
\itemsep=1pt
\item Extensive experiments demonstrate that applying AAO to existing methods results in significant performance improvements and achieves state-of-the-art results on four popular benchmarks of different tasks, i.e., AlpacaEval 2, Arena-Hard, MT-bench, Llama-Guard.
\end{itemize}

%% file: sec/method.tex
\section{Method}
\label{sec:aao}


\subsection{Preliminaries}
\label{subsec:preliminaries}



The RLHF method without a reward model starts by collecting preferred $y_w$ and dispreferred $y_l$ answers for each prompt $x$. These methods guide the LLM $\pi_{\theta}$ (initialized from the SFT model) to generate responses closer to $y_w$ and farther from $y_l$. The prompt $x$ is concatenated with $y_w$ and $y_l$ as inputs to $\pi_{\theta}$, which outputs predictions. The loss is then computed as the product of the predicted probabilities for the target tokens, as follows:
\begin{equation}
    \pi_{\theta}(y_\varepsilon |x) =\prod_{i=1}^{K_\varepsilon }  P_{\theta }(y_\varepsilon ^{(i)}|x,y_\varepsilon ^{(<i)}),
    \label{eq:pi_theta}
\end{equation}
\noindent where $\varepsilon \in \{w,l\}$, $K_{\varepsilon}$ is the number of tokens in answer $y_{\varepsilon}$, and $P_{\theta}(y_\varepsilon^{(i)}|x,y_\varepsilon^{(<i)})$ denotes the predicted likelihood for the $i^{th}$ target token in $y_\varepsilon$. The RLHF method without a reward model (e.g., DPO) aims to decrease $\pi_\theta(y_w|x)$ and increase $\pi_\theta(y_l|x)$. It also employs a reference model $\pi_{ref}$ (e.g., a frozen SFT model) to mitigate alignment deviation, using simultaneous inputs to obtain the loss factor $\pi_{ref}(y_\varepsilon|x)$. Based on these factors, this approach achieves its objective with the following function:
\begin{equation}
\begin{split}
    \mathcal{L}_{DPO}(\pi_{\theta},\pi_{ref}) = -\mathbb{E}_{(x,y_w,y_l)\sim \mathcal{D} } \\ 
    \left [ \log\sigma(\beta \log \frac{\pi_{\theta(y_w|x)}}{\pi_{ref}(y_w|x)} - \beta \log \frac{\pi_{\theta(y_l|x)}}{\pi_{ref}(y_l|x)})\right ],
\end{split}
    \label{eq:dpo_loss}
\end{equation}
\noindent where $\sigma(\cdot)$ denotes a logistic function (e.g., sigmoid). The parameter $\beta$ controls the deviation from $\pi_{ref}$. Although the specific operations of these methods are different, they all ultimately revolve around $\pi_{\theta}(y_\varepsilon|x)$ \cite{ethayarajh2024kto,azar2024general}.

\subsection{Theoretical Analysis of AAO}
\label{subsec:motivation}

Through reviewing existing alternative methods to RLHF, we note that their implementations mainly rely on the preference loss factor $\pi_{\theta}(y_\varepsilon|x)$. The mechanism of logarithmic subtraction of these factors enables LLMs to generate preferred answers. As reported in \cite{rafailov2023direct}, the gradient of the loss function $\mathcal{L}_{DPO}$ increases the likelihood of the preferred answers $y_w$ and decreases that of the dispreferred answers $y_l$. Take a close look at Eq.~\ref{eq:dpo_loss}, which can be reformulated as follows:
\begin{equation}
\begin{split}
    \mathcal{L}_{DPO}(\pi_{\theta},\pi_{ref})= -\mathbb{E}_{x,y_w,y_l\sim \mathcal{D}} \\ \left [ \log \sigma (\beta \log (\frac{\pi_{ref}(y_l|x) }{\pi_{ref}(y_w|x)}\cdot \frac{\pi_{\theta }(y_w|x) }{\pi_{\theta }(y_l|x)})) \right ], 
\end{split}
    \label{eq:dpo_2}
\end{equation}
\noindent where $\frac{\pi_{ref}(y_l|x)}{\pi_{ref}(y_w|x)}$ is a constant scaling factor due to the reference model without the need for gradient updates, while $\frac{\pi_{\theta }(y_w|x) }{\pi_{\theta }(y_l|x)}$ serves as the core component and enables LLMs to effectively achieve preference optimization. According to Eq.~\ref{eq:pi_theta}, the latter can be further transformed as follows:
\begin{equation}
\small
\label{eq:core}
\begin{aligned}
&\mathcal{L}_{core}(\pi_\theta,x,y_w,y_l) =\frac{\prod_{i=1}^{K_w}P_{\theta }(y_w^{(i)}|x,y_w^{(<i)}) }{\prod_{i=1}^{K_l}P_{\theta }(y_l^{(i)}|x,y_l^{(<i)}) } \\
&=\sum_{i=1}^{K_w} \log P_{\theta}(y_w^{(i)}|x, y_w^{(<i)}) - \sum_{j=1}^{K_l} \log P_{\theta}(y_l^{(j)}|x, y_l^{(<j)}).
\end{aligned}
\end{equation}

In the preferred answer $y_w$ and the non-preferred answer $y_l$, if there exists a common token $y^{(c)}$ appearing at position $i$ in the preferred answer and at position $j$ in the non-preferred answer, the corresponding conditional probabilities are represented as $P_{\theta}(y^{(c)} | x, y_w^{(<i)})$ and $P_{\theta}(y^{(c)} | x, y_l^{(<j)})$, respectively. Under the optimization objective of Eq.~\ref{eq:core}, the contribution term of the common token $y^{(c)}$to the log-probability difference is expressed as follows:
\begin{equation}
\log P_{\theta}(y^{(c)} | x, y_w^{(<i)}) - \log P_{\theta}(y^{(c)} | x, y_l^{(<j)})
\end{equation}

Since $y^{(c)}$ appears in different contexts in the preferred path and the non-preferred path, specifically $y_w^{(<i)} \neq y_l^{(<j)}$, the model generates conflicting gradient signals for the prediction of the same token:
\begin{itemize}
    \item  In the preferred path, the optimization pushes $\log P_{\theta}(y^{(c)} | x, y_w^{(<i)})$ to increase in order to boost the overall probability of the preferred path.
    
    \item In the non-preferred path, the optimization drives $\log P_{\theta}(y^{(c)} | x, y_l^{(<j)})$ to decrease in order to reduce the probability of the non-preferred path.
\end{itemize}

However, since$ y^{(c)}$ is the same token in both paths and shares the same model parameters, $\pi_{\theta}$ is required to simultaneously increase and decrease their predicted occurrence probabilities during optimization. As a result, their gradient signals may cancel each other out in the parameter space. Furthermore, when their contexts are similar or even identical, the gradient directions tend to be highly aligned, with almost identical magnitudes. Specifically:
\begin{equation}
\small
\nabla_{\theta} \log P_{\theta}(y^{(c)} | x, y_w^{(<i)}) \approx \nabla_{\theta} \log P_{\theta}(y^{(c)} | x, y_l^{(<j)}),
\end{equation}
their contribution to the overall gradient of the loss function approaches zero:
$\nabla_{\theta} f(\pi_{\theta}) \approx 0$

This implies that regardless of whether the contexts are entirely identical, when the same token appears in both the preferred and non-preferred answers, the optimization objective of DPO, although aiming to widen the probability gap between the preferred and non-preferred paths, struggles to effectively update the parameters at these common token positions. This optimization conflict not only could reduce convergence efficiency but also would weaken the model’s ability to distinguish between the preferred and non-preferred paths in similar contexts, thereby suppressing the contrastive learning effect of DPO.

\subsection{Identifying Ambiguous Tokens}
\label{subsec:identifying_focused_and_ambiguous_tokens}
To mitigate the negative impact of ambiguous vocabulary on preference alignment training, we propose a concise and effective method - AAO. This approach can identify ambiguous tokens in preference answers, encouraging LLMs to focus more on the core tokens that genuinely reflect human preferences. Specifically, we first encode each token in the answers into corresponding embedding vectors using the language model’s embedding layer. To determine whether a token in one answer has semantically similar or identical tokens in the other answer, we compute the cosine similarity between each token embedding in the preferred answer and all token embeddings in the rejected answer, and vice versa. We then record the maximum similarity score for each token, formulated as:
\begin{equation}
\footnotesize
S_\varepsilon^{(i)} = \max_{j \in K_{\neg \varepsilon}} \left( 
\frac{ \mathcal{F}(e_\varepsilon^{(i)}, e_{\neg \varepsilon}^{(j)}) - \min\limits_{k \in K_{\neg \varepsilon}} \mathcal{F}(e_\varepsilon^{(i)}, e_{\neg \varepsilon}^{(k)}) }
{ \max\limits_{k \in K_{\neg \varepsilon}} \mathcal{F}(e_\varepsilon^{(i)}, e_{\neg \varepsilon}^{(k)}) - \min\limits_{k \in K_{\neg \varepsilon}} \mathcal{F}(e_\varepsilon^{(i)}, e_{\neg \varepsilon}^{(k)}) } 
\right),
\label{eq:normalized_cosine_similarity}
\end{equation}
where $S_\varepsilon^{(i)}$ denotes the normalized similarity score of the i-th token in answer $y_\varepsilon$, with $\varepsilon \in \{w, l\}$ corresponding to the preferred and rejected answers, respectively. $K_\varepsilon$ is the number of tokens in the answer, and $\mathcal{F}(\cdot)$ represents the cosine similarity function between two embedding vectors. To enhance the comparability of similarity scores across different tokens, we perform min-max normalization over all cross-answer token similarities and take the normalized maximum similarity as the final score for each token. Based on this score, we introduce a threshold to distinguish different categories of tokens. The specific threshold-setting strategy is detailed in Section~\ref{subsec:adaptive_thresholds}. This approach enables the model to effectively identify and mitigate the influence of ambiguous tokens during training, thereby improving its ability to capture genuine human preference signals and enhancing preference alignment performance.

\subsection{Adaptive Thresholds}
\label{subsec:adaptive_thresholds}
Due to significant differences in semantic similarity between preferred answers across different tasks, and even among different queries, it is challenging to manually set fixed thresholds that can adapt to these varying semantic environments. To address this issue, we design a dynamic threshold adjustment mechanism that better accommodates the diverse semantics of different tasks. Specifically, we introduce a lightweight linear layer to automatically output the thresholds, formulated as follows:
\begin{equation}
  a, b = AW(P(y | x, \theta)),
\end{equation}
where $P(y | x, \theta)$ represents the output logits of the model, and $AW$ is a multi-layer perceptron (MLP). The values of parameters $a$ and $b$ are constrained within the range $[0, 1]$. To ensure validity, we introduce a clipping operation during computation to guarantee that $a > b$ is always satisfied.
This adaptive threshold module is designed to be lightweight and efficient, introducing minimal additional computational overhead while maintaining plug-and-play capability. Experimental results demonstrate that the impact of this module on time consumption is negligible (see Section~\ref{subsec:ablation}).

To further optimize the threshold outputs of the linear layer, we design an additional loss function to assist in training, enabling the model to better learn adaptive thresholds for different tasks. This loss function consists of two main components: Fine-grained Contrastive Suppression Loss and Preference Reward Enhancement Loss.

\textbf{Fine-grained contrastive suppression loss.}
To enable the reweighted generation to more precisely extract key tokens from both preferred and non-preferred data, we design a Fine-grained Contrastive Suppression Loss. Our goal is to significantly reduce the interference of background tokens and enhance the attention to key tokens, thereby optimizing the feature representation of preferred and non-preferred data more effectively.

Specifically, we define two types of similarity matrices:
\begin{equation}
\footnotesize
\begin{aligned}
&S_{\text{pref}} = \frac{\sum_{i=1}^{T_p} \sum_{j=1}^{T_d} \cos \left(E_{\text{pref}, i}\cdot w_{\text{pref,i}}, E_{\text{dis}, j}\cdot w_{\text{dis},i} \right)}{T_p \cdot T_d} ,\\
&S_{\text{dis}} = \frac{\sum_{i=1}^{T_d} \sum_{j=1}^{T_p} \cos \left(E_{\text{dis}, j}\cdot w_{\text{dis},i}, E_{\text{pref}, i}\cdot w_{\text{pref,i}} \right)}{T_d \cdot T_p} ,
\end{aligned}
\end{equation}
where $T_p$ and $T_d$ represent the number of tokens in the preferred and non-preferred data, respectively. $E_{\text{pref}}$ and $E_{\text{dis}}$ are the feature representations of the preferred and non-preferred data, while $w_{\text{pref}}$ and $w_{\text{dis}}$ are the corresponding weighting coefficients. We measure the average similarity between each token in the preferred data and all tokens in the non-preferred data, and vice versa, to quantify the feature differences between the two.

Based on this fine-grained comparison, we formulate the contrastive suppression loss as follows:
\begin{equation}
\mathcal{L}_{\text{contrastive}} = S_{\text{pref}} + S_{\text{dis}}.
\end{equation}

The optimization objective of this loss function is to maximize the difference between the preferred and non-preferred data in the high-dimensional feature space, thereby enhancing the model’s ability to capture preference expression effectively.

\textbf{Preference reward enhancement loss.}
To further improve the model’s performance on preferred samples (chosen), we introduce a Preference Reward Enhancement Loss. Unlike traditional contrastive learning, which focuses solely on the differences between positive and negative samples, our approach not only aims to increase the distinction between chosen and rejected samples but also seeks to significantly enhance the log-probability of chosen samples.

Specifically, given the model’s output probability distribution $P(y | x)$, we calculate the sum of log-probabilities over all time steps for the preferred samples as the reward metric:
\begin{equation}
R_{\text{chosen}} = \sum_{t=1}^T \log P(y_t | x, \theta),
\end{equation}
where $y_t$ represents the output of the preferred data (chosen) at time step $t$, and $\theta$ denotes the model parameters. Under the adaptive weighting mechanism, the influence of background tokens is suppressed while the significance of key tokens is magnified. This adjustment enables the model to focus more on high-information regions during the generation of preferred answers, thereby increasing the log-probability.

To further strengthen this mechanism, we introduce an optimization objective for $R_{\text{chosen}}$ in the loss function:
\begin{equation}
\mathcal{L}{\text{reward}} = -\mathbb{E}[R_{\text{chosen}}].
\end{equation}

This optimization objective implies that during each training iteration, the model is encouraged to enhance the log-probability of chosen samples, thereby amplifying the distinction between preferred and non-preferred samples in contrastive learning.

To summarize, the complete form of the auxiliary loss function is expressed as follows:
\begin{equation}
\mathcal{L}_{\text{auxiliaryloss}} = \mathcal{L}_{\text{contrastive}} + \mathcal{L}_{\text{reward}}.
\end{equation}


\subsection{Re-weighting Strategies}
\label{subsec:re_weighting_strategies}
In this study, we introduce two similarity thresholds, $a$ and $b$ (with $a > b$), to distinguish tokens based on their semantic roles. Specifically, tokens with similarity scores greater than a are classified as ambiguous tokens, which represent highly redundant tokens that may introduce confusion during training. Tokens with similarity scores below b are treated as key tokens, indicating core words that are highly discriminative and more likely to reflect true human preferences. Tokens with similarity scores between b and a are considered transitional tokens, reflecting an intermediate semantic state between ambiguous and key tokens.

After identifying these three token categories, we design a targeted reweighting strategy to suppress the influence of ambiguous noise and enhance the model’s focus on key semantics during preference alignment training. The strategy is as follows:

\textbf{Suppressing ambiguous tokens.}
For ambiguous tokens, we reduce their contribution to the training loss by down-weighting their importance based on their similarity score. As the similarity approaches 1, the weight decreases non-linearly, diminishing their impact during optimization. The weight is computed as:
\begin{equation}
   w_{\varepsilon}^{(i)} = (1 - S_\varepsilon^{(i)})^2.
\end{equation}

This effectively reduces the influence of high-redundancy tokens in the optimization process.

\textbf{Emphasizing key tokens.}
Since key tokens are more likely to carry genuine preference signals, we assign them higher training weights to encourage the model to focus on these crucial semantic units. The weighting function is defined as:
\begin{equation}
  w_{\varepsilon}^{(i)} = 1 + S_\varepsilon^{(i)}.
\end{equation}

This enhances the contribution of key tokens to parameter updates while maintaining smooth gradients.

\textbf{Preserving transitional tokens.} 
Transitional tokens have semantic similarity between ambiguous and key tokens, acting as semantic bridges to maintain information flow. During training, they are assigned a fixed weight of 1 without adjustment for several reasons: they carry auxiliary semantic details crucial for context completeness; reweighting could distort their natural distribution; keeping the weight stable balances the weakening of ambiguous tokens and emphasis on key tokens, enhancing training stability; and empirical results show limited benefits from adjusting their weights(see Section~\ref{subsec:ablation}). Thus, setting transitional tokens’ weight to 1 is a reasonable and effective choice that simplifies training while ensuring stable performance.

In summary, our final weighting formula is as follows:
\begin{equation}
w_{\varepsilon}^{(i)}=
\begin{cases}
  (1 - S_\varepsilon^{(i)})^2&     \text{ if } S_\varepsilon^{(i)} > a \\
  1&   \text{ if } a<S_\varepsilon^{(i)} < b \\
  1 + S_\varepsilon^{(i)}&   \text{ if } S_\varepsilon^{(i)} < b
\end{cases}.
\end{equation}

However, during the actual training process, traditional threshold-based decisions do not support backpropagation, resulting in ineffective parameter updates for the adaptive model. To address this issue, we redesigned an alternative formulation to overcome the limitations of non-differentiable thresholds. The new formulation is expressed as follows:
\begin{equation}
\small
\begin{aligned}
&w_{\varepsilon}^{(i)} = \frac{(1 - S_\varepsilon^{(i)})^2}{1 + e^{-\alpha \cdot ( S_\varepsilon^{(i)} - a)}}   + \frac{ (1 + S_\varepsilon^{(i)})}{1 + e^{-\alpha \cdot (b - S_\varepsilon^{(i)})}}  \\
& + (1 -  \frac{1}{1 + e^{-\alpha \cdot ( S_\varepsilon^{(i)} - a)}} - \frac{1}{1 + e^{-\alpha \cdot (b - S_\varepsilon^{(i)})}}),
\end{aligned}
\end{equation}
where $\alpha$ represents the fitting weight. When $\alpha = 200$, this parameter effectively fits Equation (16). Hence, in the subsequent experiments, we consistently set a to 200 to ensure result consistency.

Finally, we apply these weights to the logits of the final output from the LLMs, enabling the subsequent weighted training process.

%% file: sec/exp.tex
\section{Experiments and Results}
\label{sec:experiments}

\subsection{Models, Datasets, and Evaluation Metrics}
We evaluate the performance of the method from two aspects: open-domain instruction compliance benchmarks and safety alignment. During the preference optimization process, two types of models, Llama-3.1-8B~\cite{grattafiori2024llama} and Mistral-7B~\cite{jiang2024mixtral}, are evaluated in both the base setting and the instruction setting. This section aims to investigate the performance of the AAO method compared to other preference optimization approaches under different experimental conditions.

\textbf{Openended instruction-following benchmarks. }In the base setting, we follow Zephyr’s training procedure~\cite{tunstall2023zephyr} by first training a base model on the UltraChat-200k dataset~\cite{ding2023enhancing} to obtain an SFT model. Starting from this model, preference optimization is then performed on the UltraFeedback dataset~\cite{cui2023ultrafeedback}. In the instruction setting, off-the-shelf instruction-tuned models are used as the SFT models; these models have undergone more extensive instruction tuning and are more powerful and robust than the SFT models in the base setting. Preference optimization is also conducted on the UltraFeedback dataset. We evaluate method performance using three widely adopted open-domain instruction compliance benchmarks: MT-Bench~\cite{zheng2023judging}, AlpacaEval 2~\cite{alpaca_eval}, and Arena-Hard~\cite{li2024live}. 
For detailed settings, please see Table~\ref{tab:evaluation_details}.
\input{tabs/evaluation_detial}

\textbf{Safety alignment evaluation.}
In the base setting, we train a base model on the Alpaca dataset~\cite{alpaca} to obtain an SFT model, followed by preference optimization on the Anthropic-HH dataset~\cite{bai2022traininghelpfulharmlessassistant}. In the instruction setting, off-the-shelf instruction-tuned models are used as SFT models, and preference optimization is likewise performed on the Anthropic-HH dataset. 
To assess harmlessness, we generated responses using aligned LLM on a mixed dataset of AdvBench~\cite{zou2023universal} and JailbreakBench~\cite{chao2024jailbreakbench}, and used Llama-Guard~\cite{inan2023llama} to determine the safety of the responses.

\input{tabs/compared_result}

\subsection{Baseline Methods and Training Settings}
We conducted comparative experiments to evaluate the proposed method against various baseline alignment methods, including sequence-level approaches such as DPO~\cite{rafailov2023direct}, IPO~\cite{azar2024general}, KTO~\cite{ethayarajh2024kto}, and SimPO~\cite{meng2024simpo}, as well as token-level methods such as TDPO~\cite{zeng2024token}, RTO~\cite{zhong2024dpo}, and TIS-DPO~\cite{liu2024tis}. In addition, we introduced a randomly weighted method, DPO-Random, to further validate the effectiveness of our proposed weighting strategy. All baseline methods use the hyperparameter settings provided in their original papers. For our method, the learning rate is set to 5e-7, the batch size to 16, and training is performed for one epoch using the AdamW optimizer.

\subsection{Main results}
Table~\ref{tab:results} presents our main experimental results. Despite its simplicity, our method achieves significant improvements across all metrics. On the LLaMA-3.1-8B-Base model, it raises the WR of AlpacaEval2 to 40.23\%, 7.2 points higher than the second-best method. In the Arena-Hard benchmark, it further boosts performance to 41\%, validating its effectiveness in enhancing generalization.
Although MT-Bench is widely used, it shows weak differentiation across methods due to its limited data size and single-instance scoring~\cite{meng2024simpo, li2024live}. By contrast, AlpacaEval2 and Arena-Hard provide more reliable assessments. Even so, our method still achieves top performance across models and tasks, demonstrating strong robustness and adaptability.
For safety alignment, our method scores 91.29\% on LLaMA-3.1-8B-Base, 3.42 points above the second-best, indicating better distinction between preferred and non-preferred data.

\subsection{Ablation Study}
\label{subsec:ablation}
\begin{figure}[t]
  \centering
  \includegraphics[trim=0 45 350 0, clip,width=0.35\textwidth]{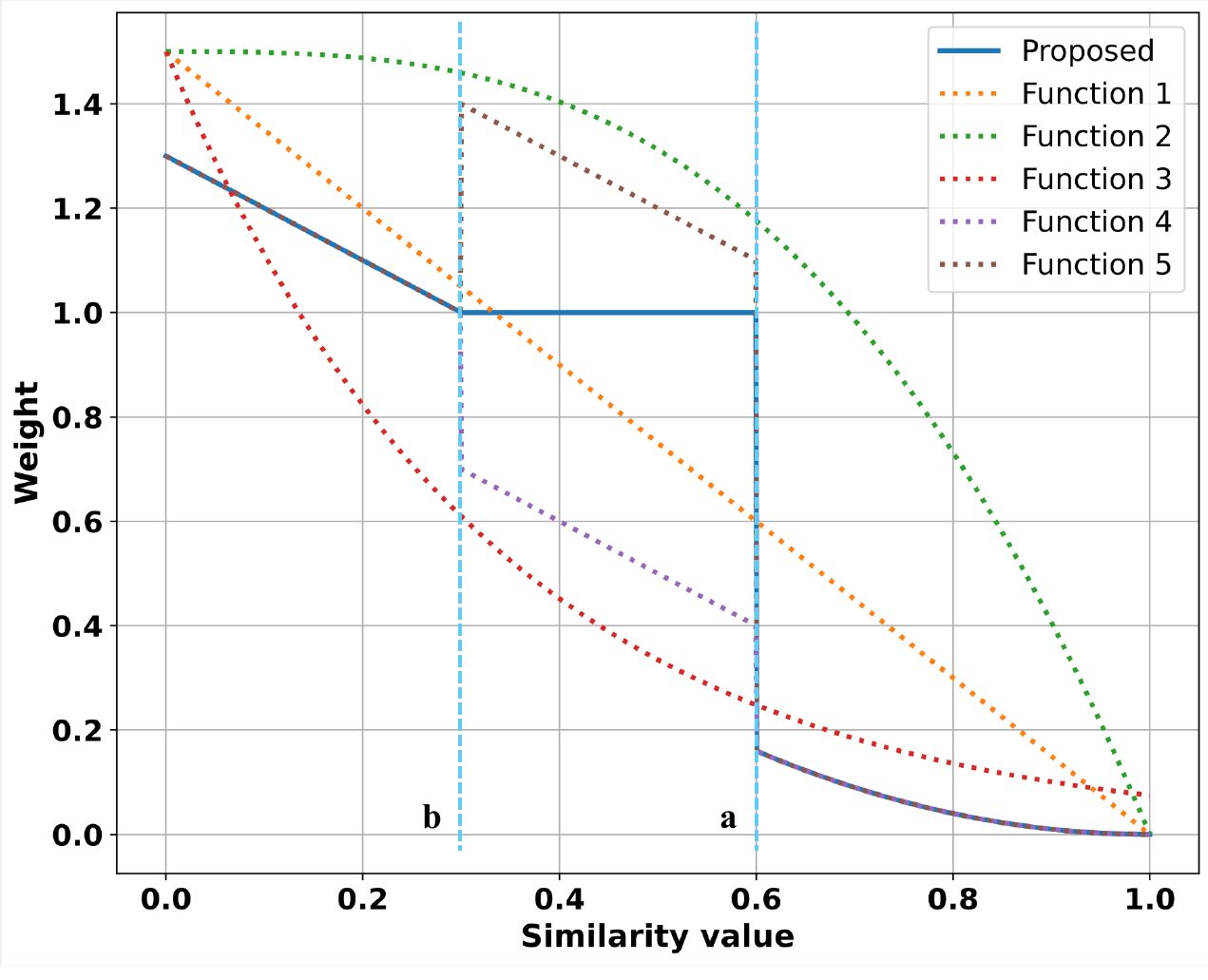}
  \caption{Images of different weighted curves. In our approach, the thresholds $a$ and $b$ are decided by LLM itself during training.}
  \label{fig:curves}
  \vspace{-1em}
\end{figure}

\textbf{Analysis of weighting function.}
To validate the effectiveness of the proposed weighting strategy, we designed multiple weighting curves with different trends for comparative experiments. Among them, Functions (4) and (5) perform differentiated weighting operations on transitional tokens based on our strategy. The specific trends of each weighting curve are illustrated in Figure~\ref{fig:curves}, and the corresponding experimental results are presented in Table~\ref{tab:curves}.
The experimental results clearly demonstrate that the proposed adaptive weighting strategy exhibits significant advantages across all metrics, fully proving its superiority. Moreover, when the weights of transitional tokens are adjusted, the model’s training process is significantly affected, further validating the correctness of our hypothesis. Notably, the design of most weighting curves effectively optimizes the training performance of DPO, which also reveals that the ambiguous phenomenon we proposed is both reasonable and existent.
It should be noted that the choice of weighting curves remains an open question. 
In this study, we demonstrated the significant optimization effects of the proposed weighting curves, and future research could further explore more weighting strategies to enhance model performance.

\input{tabs/curves}

\begin{figure}[t]
  \centering
  \includegraphics[width=0.4\textwidth]{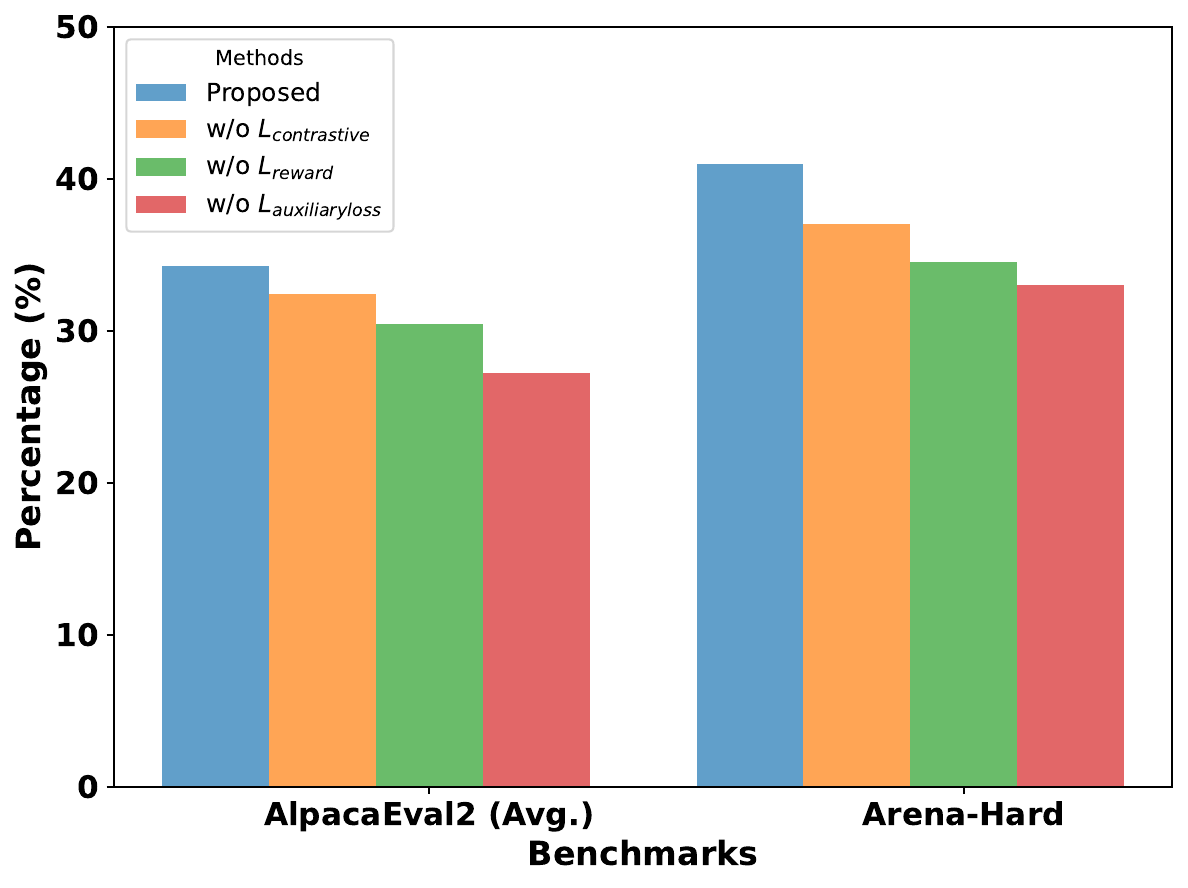}
  \caption{Effect of auxiliary losses on experimental results.}
  \label{fig:auxiliary}
  \vspace{-1.5em}
\end{figure}

\textbf{Analysis of the auxiliary loss function.}
We further explored the impact of each component in the auxiliary loss function on the final results, as illustrated in Figure~\ref{fig:auxiliary}. The results indicate that different components exhibit varying degrees of influence on the optimization of the adaptive module. When these components work synergistically, the model achieves optimal performance, validating the effectiveness and rationality of our design.

\textbf{Analysis of time consumption.}
We conducted alignment experiments based on the Mistral-7B-Base model on the UltraFeedback dataset and compared the time overhead between DPO and AAO during the training process. The experimental results are presented in Table~\ref{tab:consumption}. The results indicate that our proposed adaptive model only consumes an additional 62.50 MB of storage space while not significantly increasing the training time, demonstrating its efficiency and lightweight design.

\input{tabs/consumption}

\begin{figure}[t]
  \centering
  \includegraphics[width=0.4\textwidth]{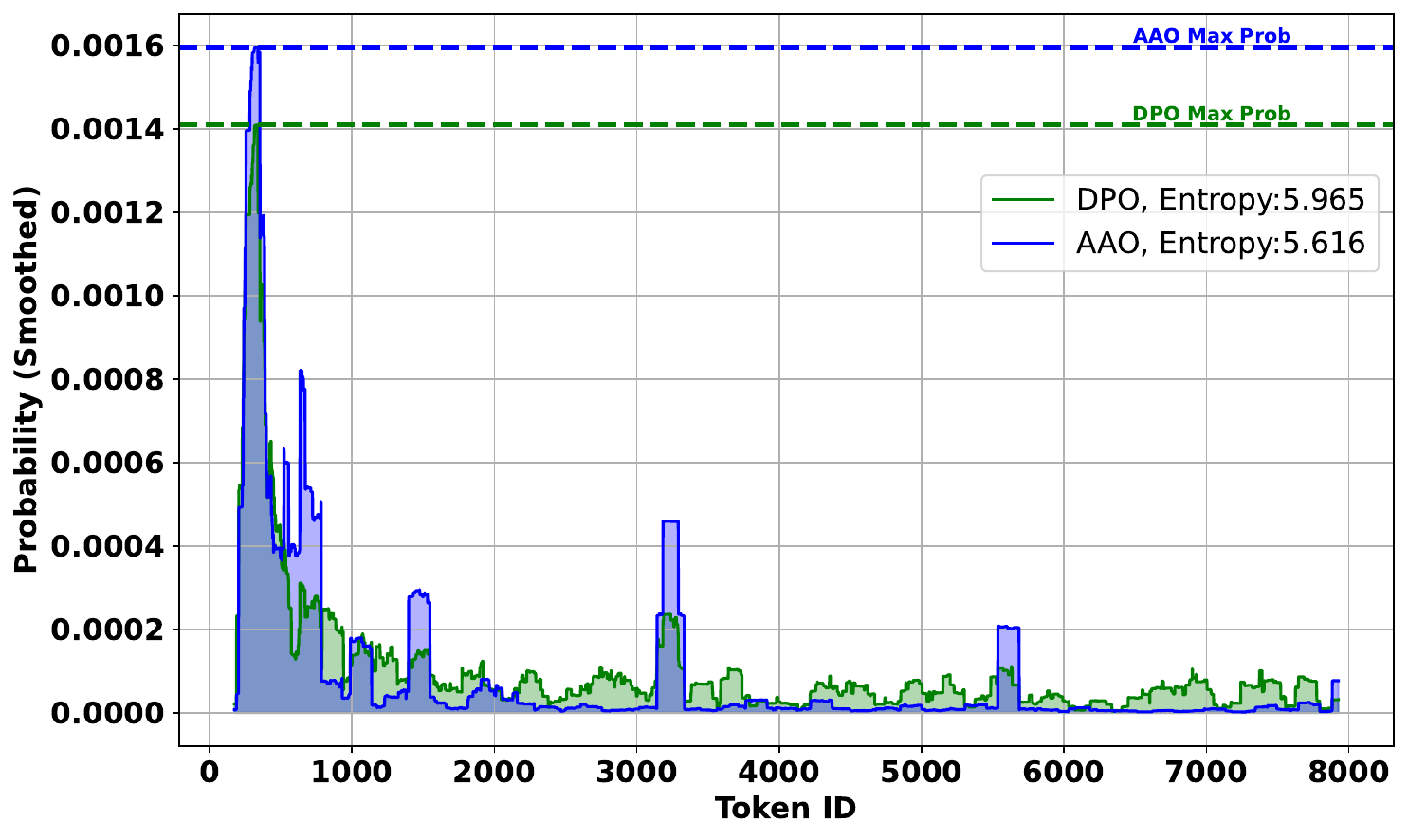}
  \caption{AAO mitigates the squeeze effect of DPO.}
  \label{fig:squeeze}
  \vspace{-1em}
\end{figure}

\textbf{Squeeze effect.}
We further validated the effectiveness of AAO in alleviating the “squeezing effect” observed in DPO \cite{ren2024learning}. As illustrated in Figure~\ref{fig:squeeze}, during the training process, DPO significantly reduces the confidence of the highest-probability token while correspondingly increasing the probabilities of other tokens, clearly revealing the presence of the “squeezing effect.” In contrast, our method effectively mitigates this phenomenon, further indicating that the cause of the “squeezing effect” is not solely attributed to sequence-level semantic similarity but also involves the handling of internal ambiguous labels.

%% file: tabs/evaluation_detial.tex

\begin{table}[]
\setlength{\tabcolsep}{1.5pt}
\centering
\footnotesize
\begin{tabular}{cccc}
\hline
                 & Baseline & Judge Model & Metric             \\ \hline
AlpacaEval 2        & GPT-4 Turbo    & GPT-4.1     & LC \& raw win rate \\
Arena-Hard         & GPT-4-0314         & GPT-4o     & Win rate           \\
MT-Bench            & -              & GPT-4o     & Rating of 1-10     \\ \hline
\end{tabular}
\caption{Evaluation details for AlpacaEval 2, Arena-Hard, and MT-Bench. The baseline model refers to the model compared against.}
\label{tab:evaluation_details}
\vspace{-1em}
\end{table}

%% file: tabs/compared_result.tex
\begin{table*}[]
\centering
\setlength{\tabcolsep}{0.6pt}
\footnotesize
\begin{tabular}{cccccc|ccccc}
\hline
                         & \multicolumn{2}{c}{AlpacaEval2} & MT-Bench & Arena-Hard & Llama-Guard & \multicolumn{2}{c}{AlpacaEval2} & MT-Bench & Arena-Hard & Llama-Guard \\ \cline{2-11} 
\multirow{-2}{*}{Method} & LC(\%)↑        & WR(\%)↑        & Avg.↑    & WR(\%)↑    & Harm.(\%)↑  & LC(\%)↑        & WR(\%)↑        & Avg.↑    & WR(\%)↑    & Harm.(\%)↑  \\ \hline
                         & \multicolumn{5}{c|}{\cellcolor[HTML]{CBCEFB}Llama3.1-8B-Base}         & \multicolumn{5}{c}{\cellcolor[HTML]{CBCEFB}Llama3.1-8B-Instruct}      \\ \hline
DPO                       & 22.45          & 31.30          & 7.36    & 26.0      & 83.06       & 45.87          & 44.34          & 7.73    & 29.6      & 96.12       \\
IPO                      & 24.38          & 24.90          & 7.41    & 25.1      & 82.14       & 45.43          & 44.56          & 7.82    & 28.3      & 97.42       \\
KTO                      & 25.79          & 24.79          & 7.41    & 25.8      & 83.42       & 43.86          & 42.00          & 7.86    & 26.8      & 96.52       \\
SimPO                    &27.45          & 33.03          & 7.47    & 30.6    & 84.15      & 47.50          & 43.64           & 7.91    & 33.5      & 97.85       \\
TDPO                     & 23.34          & 26.45          & 7.22    & 27.2      & 86.54       & 46.56          & 43.21          & 7.75    & 26.4      & 97.04       \\
RTO                      & 24.43          & 25.84          & 7.34    & 26.7      & 85.71       & 46.84          & 41.98          & 7.81    & 30.4      & 96.99       \\
TIS-DPO                  & 26.84          & 31.47          & 7.40    & 27.8      & 87.87       & 47.04         & 43.99          & 7.86    & 30.1      & 97.54       \\
DPO-Random                  & 19.07          & 33.97          & 7.11    & 25.4      & 83.42       & 36.36         & 35.10          & 7.33    & 27.9      & 97.32       \\
\hline
AAO                      & \textbf{28.36}          & \textbf{40.23}          & \textbf{7.52}    & \textbf{41.0}      & \textbf{91.29}       & \textbf{48.11}          & \textbf{44.89}          & \textbf{7.92}    & \textbf{42.7}      & \textbf{98.42} 
      \\ \hline
                         & \multicolumn{5}{c|}{\cellcolor[HTML]{CBCEFB}Mistral-7B-Base}          & \multicolumn{5}{c}{\cellcolor[HTML]{CBCEFB}Mistral-7B-Instruct}       \\ \hline
DPO                      & 20.45          & 17.55          & 7.16    & 8.1      & 95.48       & 32.62          & 28.70          & 7.51    & 17.0      & 99.14       \\
IPO                      & 17.13          & 12.13          & 7.33    & 7.5      & 96.34       & 31.46          & 30.31          & 7.56    & 16.8      & 99.42       \\
KTO                      & 16.49          & 09.32          & 7.31    & 8.2      & 95.04       & 34.65          & 30.90          & 7.64    & 18.6      & 99.21       \\
SimPO                    & 24.40          & 22.80          & 7.34    & 9.7      & 96.56       & 35.93          & 34.48          & 7.62    & 22.6      & 99.35       \\
TDPO                     & 21.47          & 18.56          & 7.18    & 7.9      & 94.55       & 35.34          & 30.55          & 7.52    & 18.3      & 97.97       \\
RTO                      & 20.40          & 16.77          & 7.26    & 8.1      & 94.36       & 34.49          & 32.52          & 7.55    & 17.9      & 98.76       \\
TIS-DPO                  & 21.57          & 18.84          & 7.25    & 8.2      & 96.74       & 34.54          & 33.51          & 7.58    & 18.0      & 97.35       \\
DPO-Random                  & 17.71          & 11.22          & 7.14    & 8.6      & 95.52       & 30.14         & 28.45          & 7.50    & 16.8      & 98.42       \\
\hline
AAO                      & \textbf{25.01}          & \textbf{23.45}          & \textbf{7.36}    & \textbf{13.6}      & \textbf{97.41}       & \textbf{36.66}          & \textbf{38.54}          & \textbf{7.65}    & \textbf{24.2}      & \textbf{99.67}
      \\ \hline
\end{tabular}
\caption{AlpacaEval 2, Arena-Hard, MT-Bench and Llama-Guard results under the four settings. LC, WR, and Harm. represent length-controlled, raw win rate, and harmless response rate, respectively.}
\label{tab:results}
\end{table*}

%% file: tabs/curves.tex
\begin{table}[]
\normalsize
\centering
\footnotesize
\setlength{\tabcolsep}{1.5pt}
\begin{tabular}{ccccc}
\hline
\multirow{2}{*}{Method} & \multicolumn{2}{c}{AlpacaEval2} & MT-Bench & Arena-Hard \\
                        & LC(\%)↑        & WR(\%)↑        & Avg.↑    & WR(\%)↑    \\ \hline
Function 1              & 22.68          & 28.52          & 6.94    & 29.45      \\
Function 2              & 23.74          & 32.38          & 7.14    & 32.44      \\
Function 3              & 18.56          & 14.51          & 6.87    & 24.69      \\ 
Function 4              & 27.98          & 37.24          & 7.42    & 38.84      \\ 
Function 5              & 27.32          & 35.17          & 7.47    & 37.45      \\ 
\hline
Proposed                &\textbf{28.36}          & \textbf{40.23}          & \textbf{7.52}    & \textbf{41.00}     \\ \hline
\end{tabular}
\caption{Results under different weighting methods.}
\label{tab:curves}
\end{table}

%% file: tabs/consumption.tex
\begin{table}[]
\centering
\footnotesize
\begin{tabular}{ccc}
\hline
    & Extra parameter size & Training time \\ \hline
DPO & -                    &       77.81 mins        \\
AAO &        62.50MB              &   78.12 mins (0.4\% ↑)            \\ \hline
\end{tabular}
\caption{Comparison of training time and additional parameters.}
\label{tab:consumption}
\vspace{-1em}
\end{table}

%% file: sec/related_work.tex
\section{Related Work}
\label{sec:related_work}
RLHF has become a leading approach for aligning LLMs with human values, typically involving three stages: supervised fine-tuning, reward model training, and policy optimization using algorithms like PPO, GRPO, and REINFORCE++ \cite{ouyang2022training, casper2023open, achiam2023gpt}. Despite its effectiveness, RLHF’s reliance on multiple models and LLM sampling increases complexity \cite{zhao2023slic, casper2023open}. To address this, efficient offline alternatives like DPO have emerged, which transfer preference knowledge using both preferred and non-preferred responses without explicit reward models \cite{rafailov2023direct, xu2024contrastive, meng2024simpo}.


Recent studies indicate DPO optimizes LLMs by considering whole-response preferences, ignoring token-level importance, which constrains its performance \cite{zeng2024token,liu2024tis,gu2025mask}. To address this, our work explores the impact of token importance in DPO training, observing that semantically similar tokens in positive and negative examples may cause confusion. 
To solve this, we introduce AAO, an approach that allows the model to re-weight tokens automatically, mitigating confusion and enhancing DPO performance. Our analysis further uncovers that semantic confusion may contribute to the “squeeze effect” identified in \cite{ren2024learning}.

%% file: sec/con.tex
\section{Conclusion}
In this paper, we theoretically and empirically identify potential ambiguity issues in DPO. To address this problem, we propose AAO, a simple yet effective approach that automatically re-weights background content based on semantic similarity. Extensive experiments demonstrate that AAO consistently outperforms existing methods across various training setups, validating its effectiveness. 

%% file: sec/limitations.tex
\section{limitations}
\textbf{Rigorous theoretical analysis.} Although AAO has achieved practical success and is intuitively motivated, a more rigorous theoretical and experimental analysis is still necessary to fully understand the impact of ambiguous content during DPO training. Ideally, this can be accomplished by tracking the gradient dynamics associated with ambiguous content and combining these observations with theoretical derivations, in order to obtain more definitive conclusions and insights. We will leave this aspect for future work, aiming to gain a deeper understanding and provide a more quantitative analysis of the impact of ambiguous tokens during training.

\textbf{Design of re-weighting strategy.} 
In this study, we propose a piecewise similarity-based reweighting curve, designed based on empirical assumptions, to mitigate the adverse effects of ambiguous tokens. Our approach outperforms other curve designs and achieves state-of-the-art results. However, it may not be optimal; future work could explore enabling the model to automatically fit such curves or investigate novel mapping strategies.